\documentclass[runningheads]{llncs}

\usepackage{iciap}

\usepackage{iciapabbrv}

\usepackage{graphicx}
\usepackage{booktabs}

\usepackage[accsupp]{axessibility}  % Improves PDF readability for those with disabilities.

\usepackage[pagebackref,breaklinks,colorlinks,citecolor=iciapblue]{hyperref}
\usepackage{orcidlink}

\begin{document}

\title{WhoFi: Deep Person Re-Identification\\
 via Wi-Fi Channel Signal Encoding} 

\titlerunning{WhoFi}

\author{Danilo Avola \and Emad Emam \and Dario Montagnini \and Daniele Pannone \and \\Amedeo Ranaldi  }

% Author1\inst{2,3}\orcidlink{1111-2222-3333-4444} \and
% Author2\inst{3}\orcidlink{2222--3333-4444-5555}

\authorrunning{D. Avola et al.}

\institute{Department of Computer Science, La Sapienza University of Rome \\
\email{\{avola, emam, montagnini, pannone, ranaldi\}@di.uniroma1.it}\\
}

\maketitle

\begin{abstract}
Person Re-Identification is a key and challenging task in video surveillance. While traditional methods rely on visual data, issues like poor lighting, occlusion, and suboptimal angles often hinder performance. To address these challenges, we introduce WhoFi, a novel pipeline that utilizes Wi-Fi signals for person re-identification. Biometric features are extracted from Channel State Information (CSI) and processed through a modular Deep Neural Network (DNN) featuring a Transformer-based encoder. The network is trained using an in-batch negative loss function to learn robust and generalizable biometric signatures. Experiments on the NTU-Fi dataset show that our approach achieves competitive results compared to state-of-the-art methods, confirming its effectiveness in identifying individuals via Wi-Fi signals.
\keywords{Person Re-Identification \and CSI \and Deep Neural Networks \and Transformers \and Wi-Fi Signals \and Radio Biometric Signature}
\end{abstract}

\section{Introduction}
\label{sec:intro}
Person Re-Identification (Re-ID) plays a central role in surveillance systems, aiming to determine whether two representations belong to the same individual across different times or locations. Traditional Re-ID systems typically rely on visual data such as images or videos, comparing a probe (the input to be identified) against a set of stored gallery samples by learning discriminative biometric features. Most commonly, these features are based on appearance cues such as clothing texture, color, and body shape. However, visual-based systems suffer from a number of known limitations, including sensitivity to changes in lighting conditions~\cite{seco}, occlusions~\cite{occ}, background clutter~\cite{cltr}, and variations in camera viewpoints~\cite{prim}. These challenges often result in reduced robustness, especially in unconstrained or real-world environments. To overcome these limitations, an alternative research direction explores non-visual modalities, such as Wi-Fi-based person Re-ID. Wi-Fi signals offer several advantages over camera-based approaches: they are not affected by illumination, they can penetrate walls and occlusions, and most importantly, they offer a privacy-preserving mechanism for sensing. The core insight is that as a Wi-Fi signal propagates through an environment, its waveform is altered by the presence and physical characteristics of objects and people along its path. These alterations, captured in the form of Channel State Information (CSI), contain rich biometric information. Unlike optical systems that perceive only the outer surface of a person, Wi-Fi signals interact with internal structures, such as bones, organs, and body composition, resulting in person-specific signal distortions that act as a unique signature.

Earlier wireless sensing methods primarily relied on coarse signal measurements such as the Received Signal Strength Indicator (RSSI)~\cite{OMH14}, which proved insufficient for fine-grained recognition tasks. More recently, CSI has emerged as a powerful alternative~\cite{YZL13}. CSI provides subcarrier-level measurements across multiple antennas and frequencies, enabling a detailed and time-resolved view of how radio signals interact with the human body and surrounding environment. By learning patterns from CSI sequences, it is possible to perform Re-ID by capturing and matching these radio biometric signatures. Despite the promising nature of Wi-Fi-based Re-ID, the field remains underexplored, especially in terms of developing scalable deep learning methods that can generalize across individuals and sensing environments. In this paper, we propose WhoFi, a deep learning pipeline for person Re-ID using only CSI data. Our model is trained with an in-batch negative loss to learn robust embeddings from CSI sequences. We evaluate multiple backbone architectures for sequence modeling, including Long Short-Term Memory (LSTM), Bidirectional LSTM (Bi-LSTM), and Transformer networks, each designed to capture temporal dependencies and contextual patterns. The main contributions of this work are:
\begin{itemize}
    \item We propose a modular deep learning pipeline for person Re-ID that relies solely on Wi-Fi CSI data, without requiring visual input;
    \item We perform a comparative study across three widely used backbone architectures (LSTM, Bi-LSTM, and Transformer networks) to assess their ability to encode biometric signatures from CSI;
    \item We adopt an in-batch negative loss training strategy, which enables scalable and effective similarity learning in the absence of labeled pairs;
    \item We conduct extensive experiments on the public NTU-Fi dataset to demonstrate the accuracy and generalizability of our approach;
    \item We perform an ablation study to evaluate the impact of preprocessing strategies, input sequence length, model depth, and data augmentation.
\end{itemize}

By leveraging non-visual biometric features embedded in Wi-Fi CSI, this study offers a privacy-preserving and robust approach for Wi-Fi-based Re-ID, and it lays the foundation for future work in wireless biometric sensing.

\section{Related Work}  
\subsection{Person Re-Identification via Visual Data}
In the field of computer vision, person Re-ID has long been of major importance. Earlier methods primarily relied on RGB images or videos to track people across camera views. Handcrafted descriptors such as Local Binary Patterns (LBP), color histograms, and Histograms of Oriented Gradients (HOG) were widely used to capture low-level visual cues like texture and silhouette. With the advent of deep learning, Convolutional Neural Networks (CNNs) became the dominant approach, enabling hierarchical spatial feature learning~\cite{jml}. Training strategies like triplet loss, cross-entropy with label smoothing, and center loss were adopted to optimize embedding space separability~\cite{hermans2017, zheng19}. Recent models often integrate attention mechanisms~\cite{li} and part-based representations~\cite{sun2018} to handle misalignment and occlusion. Despite strong benchmark performance, these systems rely heavily on high-quality visual input and careful manual tuning, limiting their applicability in uncontrolled environments.
\subsection{Person Identification and Re-ID via Wi-Fi Sensing}
Several works have extensively investigated human identification and authentication through Wi-Fi CSI, focusing on features such as amplitude, phase, and heatmap variations~\cite{dua}. Early methods include line-of-sight waveform modeling combined with PCA or DWT for classification~\cite{xin2016}, or gait-based identification through handcrafted features~\cite{zpm16}. CAUTION~\cite{dataset2} introduced a dataset and few-shot learning approach for user recognition via downsampled CSI representations. More recent methods leverage deep learning models to enhance generalization capabilities~\cite{dataset1}. A recent approach~\cite{avola} proposed a dual-branch architecture that combines CNN-based processing of amplitude-derived heatmaps with LSTM-based modeling of phase information for re-identification. However, the use of private datasets in such work limits replicability and hinders direct comparison. In contrast, our study relies on a widely available public benchmark, enabling reproducibility and fair evaluation across different architectures.

\section{Method}
In this section, details about data pre-processing and augmentation, together with the proposed deep architecture, are presented.
\subsection{Data Pre-processing}
Data extracted from the CSI complex matrix must first be pre-processed to remove noise and sampling offsets to extract meaningful biometric features.
\subsubsection{Channel State Information (CSI):}
Wi-Fi transmission relies on electromagnetic waves that carry information from a transmitting antenna (TX) to a receiving one (RX). Modern systems adopt Multiple-Input Multiple-Output (MIMO), involving multiple TX/RX antennas, and Orthogonal Frequency-Division Multiplexing (OFDM), a modulation technique that transmits data across orthogonal subcarriers spanning nearly the entire frequency band. The integration of MIMO and OFDM enables sampling of the Channel Frequency Response (CFR) at subcarrier granularity in a CSI matrix. The CSI measurement for each subcarrier $k \in K$ represents the CFR $H^{(\theta,\gamma)}$ between the receiving antenna (RX) $\theta \in \Theta$ and the transmitting antenna (TX) $\gamma \in \Gamma$, and is given by:
\begin{equation}
    H^{(\theta,\gamma)}_{k} = |H^{(\theta,\gamma)}_{k}| e^{j \angle H^{(\theta,\gamma)}_{k}},
    \label{eq:CSI}
\end{equation}
where $|H^{(\theta,\gamma)}_{k}|$ denotes the signal amplitude and $\angle H^{(\theta,\gamma)}_{k}$ the signal phase. By collecting the responses across all TX/RX antenna pairs, a CSI complex matrix of size $\Theta \times \Gamma \times K$ is formed, representing the CFR across all subcarriers in $K$.
\subsubsection{Amplitude Filtering:}
Signal amplitude represents the strength of the received signal. For a subcarrier $k \in K$, receiver antenna $\theta \in \Theta$, and transmitter antenna $\gamma \in \Gamma$, the signal amplitude $A^{(\theta,\gamma)}_{k}$ is defined as:
\begin{equation}
    A^{(\theta,\gamma)}_{k} = |H^{(\theta,\gamma)}_{k}| = \sqrt{\text{real}(H^{(\theta,\gamma)}_{k})^2 + \text{img}(H^{(\theta,\gamma)}_{k})^2},
    \label{eq:amp_eq}
\end{equation}
which corresponds to the magnitude of the CSI measurement. In this work, signal amplitudes are cleaned of outliers using the Hampel filter~\cite{davies1993identification}, which identifies outliers based on the median of a local window and the Median Absolute Deviation (MAD). Given a sequence of amplitude values across $p$ packets, the local window $W^{p,k}$ of size $w$ (set to 5) centered on packet $p$ is defined as:
\begin{equation}
W^{p,k} = \left\{ A^{p - \lfloor w/2 \rfloor}_k, \ldots, A^{p + \lfloor w/2 \rfloor}_k \;:\; A^{p - \lfloor w/2 \rfloor}_k < A^{p + \lfloor w/2 \rfloor}_k \right\},
\end{equation}
\begin{equation}
\text{median}(W^{p,k}) = W^{(p,k)}_{\left\lfloor w/2 \right\rfloor},
\end{equation}
\begin{equation}
\text{MAD}(W^{p,k}) = \text{median}(|W^{p,k}_{i} - \text{median}(W^{p,k})|) \quad \forall i, \; 1 \le i \le w,
\end{equation}
where $W^{p,k}$ denotes the vector containing the $w$ neighboring data packets centered at packet $p$, sorted in ascending order, for the $k$-th subcarrier. An amplitude value is classified as an outlier if its deviation from the local median exceeds a fixed threshold. Specifically, any value outside the range:
\begin{equation}
    \text{limit}_{p,k} = \text{median}(W^{p,k}) \pm \xi \cdot \text{MAD}(W^{p,k}),
    \label{eq:mad_calc}
\end{equation}
with $\xi$ set to 3, is considered an outlier and removed.
\subsubsection{Phase Sanitization:}
Signal phase represents the temporal shift of a signal. It is calculated as the arctangent of the imaginary and real parts of the CFR:
\begin{equation}
    P^{(\theta,\gamma)}_{k} = \tan^{-1} \left( \frac{\text{img} (H^{(\theta,\gamma)}_{k})}{\text{real}(H^{(\theta,\gamma)}_{k})}\right).
\end{equation}
To remove any possible phase shifts caused by imperfect synchronization between the transmitter and receiver hardware components, we apply a standard linear phase sanitization technique. The estimated phase $\angle \hat{H}(f)_{k}$ at frequency $f$ from the CSI measurements is expressed as:
\begin{equation}
    \angle \hat{H}(f)_{k} = H(f)_{k} + 2\pi \frac{m_k}{N} \Delta t + \beta + Z,
\end{equation}
where, $H(f)_{k}$ is the actual phase, $\Delta t$ is a time offset from any delay in the signal arrival and reception, $\beta$ is the unknown phase offset, and $Z$ is a noise factor. Since the delay factor is a linear function in the subcarrier index $m_k$, it is possible to estimate the correct phase slope $a$ and offset $b$ as:
\begin{equation}
a = \frac{\angle \hat{H}(f)_{K} - \angle \hat{H}(f)_{1}}{m_K - m_1},
\end{equation}
\begin{equation} 
b = \frac{1}{K} \sum_{k=1}^{K} \angle \hat{H}(f)_{k}.
\end{equation}
Therefore, the calibrated phase $\angle H'(f)_{k}$ for each subcarrier $k \in K$ can be estimated by subtracting a linear term from the raw phase as:
\begin{equation}
     \angle H'(f)_{k} = \angle \hat{H}(f)_{k} - a m_k + b.
\end{equation}
\subsection{Data Augmentation}
To enhance model sensitivity and overall robustness against noise or minor signal fluctuations, we apply several data augmentation techniques during training. These transformations are performed on the extracted amplitude features rather than directly on the raw CSI data. For each amplitude entry, one augmentation is applied with a 90\% probability, leaving the remaining 10\% unmodified. The first augmentation adds Gaussian noise $n(t) \sim \mathcal{N}(0, \sigma^2)$ to the amplitude value $A^{(\theta,\gamma)}_k(t)$ at each time step $t$, where $\sigma = 0.02$, simulating realistic signal fluctuations and improving generalization in noisy environments. The second augmentation scales the amplitude by a random factor uniformly sampled in $[0.9, 1.1]$, modeling small variations in signal strength due to environmental or device-related factors. Finally, a time shift is applied by offsetting the amplitude sequence forward or backward by a random integer $t' \in [-5, 5]$ within a sequence of length $P = 100$. Any value shifted outside the sequence bounds is replaced with the mean amplitude of the original signal, simulating delays or de-synchronizations in signal acquisition.
\subsection{Deep Neural Network Architecture}
\begin{figure}[t]
    \centering
\includegraphics[width=0.8\textwidth]{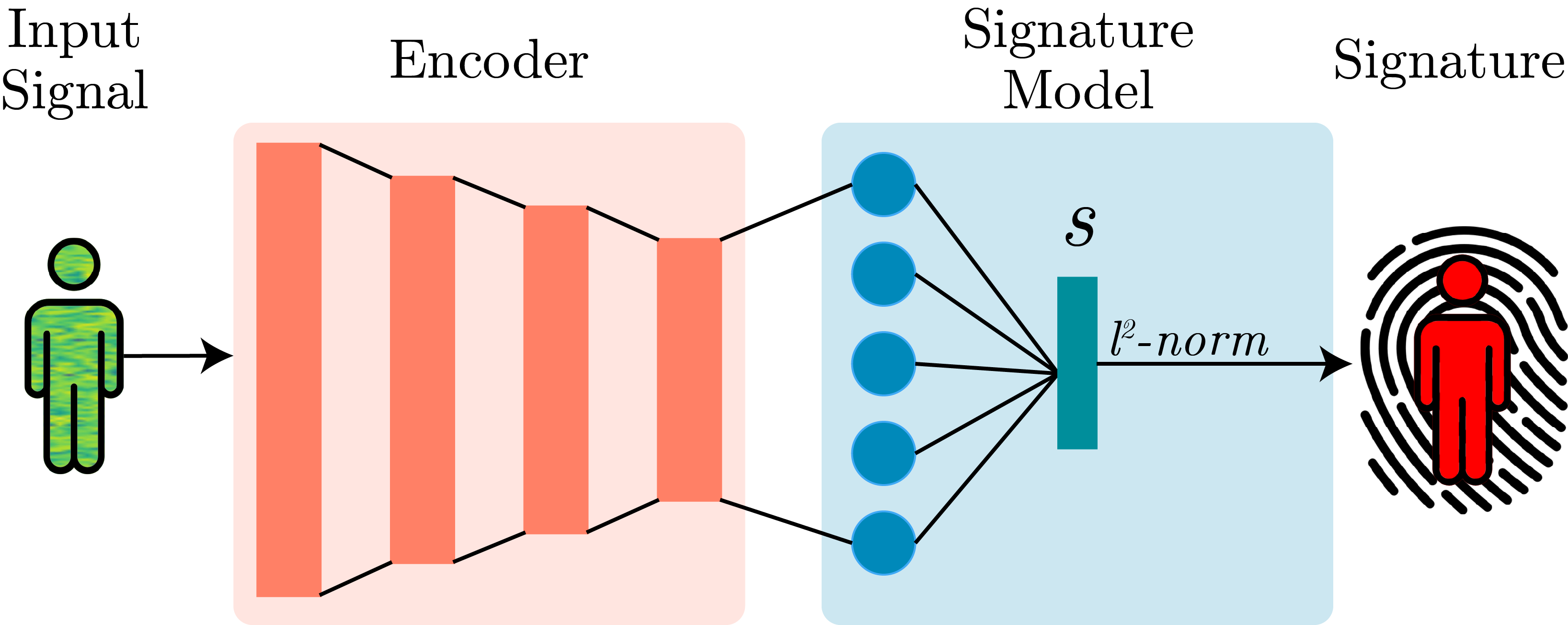}
    \caption{Overview of the proposed framework. The system takes an input signal (e.g., a person sensing data) and processes it through an encoder that extracts meaningful latent representations. These features are passed to a signature model that computes a compact signature vector $s$. To ensure consistency and comparability, the output signature is normalized through the $l^2$ normalization. The resulting signature serves as a unique identifier for the individual based on the input signal characteristics.}
    \label{fig:DNN}
\end{figure}

In the proposed pipeline, a DNN is designed to generate a biometric signature from the processed CSI features. The architecture is composed of an Encoder module ($M_e$) and a Signature Module ($M_s$) as shown in Figure \ref{fig:DNN}.
\subsubsection{Encoder Module:}
The encoder module produces a fixed-size vector that contains human signature relevant information from the provided CSI measurements. This module aims at extracting low-dimensional encoding of the high-dimensional and sequential inputs, including the amplitude or phase extracted from the CSI measurement of the wireless channel while a specific person is present between the transmitter and receiver. This work evaluates three types of encoding architectures compatible with sequential data: an LSTM encoder, a Bi-LSTM encoder, and the encoder part of a Transformer model:
\begin{enumerate}
    \item \textbf{LSTM Encoder:} LSTMs capture temporal dependencies in input sequences, enabling the model to recognize recurrent patterns. The LSTM encoder consists of $l$ stacked hidden units, where the output of the $l_i$-th unit is passed to the $l_{i+1}$-th unit in the hidden layer. Dropout layers with probability $p_d$ are interleaved between LSTM layers to improve robustness and reduce overfitting during training. The final hidden state $H^l$ from the last LSTM layer serves as the encoded output.
    \item \textbf{Bi-LSTM Encoder:} Bi-LSTMs are able to capture the correlation between time steps in the input sequence by processing the sequence in both forward and direction. This allows the model to capture context from both past and future time steps. Similar to the LSTM encoder, $l$ stacked BiLSTM layers with interleaved dropout layers are used to avoid overfitting. The last hidden states from both forward ($\overrightarrow{H^l}$) and backward ($\overleftarrow{H^l}$) passes are concatenated to form the output encoding $H^l$.
    \item \textbf{Transformer Encoder:} The encoder from the Transformer architecture is capable of detecting correlation between different elements in distant time steps in the input sequence. The encoder contains $l$ identical layers, each containing a multi-head self-attention sub-layer and a position-wise feed-forward network sub-layer. Standard and non-trainable sinusoidal positional encodings are added to the input embeddings to retain sequence order information.  Moreover, residual connections and layer normalization are applied after each sub-layer. A Dropout layer with drop probability $p_d$ is used in-between encoder layers as a regularization technique. The output of the final Transformer layer acts as the encoded representation.
\end{enumerate}
\subsubsection{Signature Module:}
The Signature module takes the fixed-size vector output from the encoder module and generates a final biometric signature. It consists of a linear layer and a $l^2$ normalization function. The linear layer is a fully connected layer that maps the encoder output to the desired signature $s$-dimensional space. Then, a normalization function is applied to regularize and uniform the output vector values to have a unit $l^2$ norm. Therefore, normalization ensures that the signatures lie on a hypersphere, which facilitates the similarity computations used in the loss function, thus, speeding up the training phase. 
\subsection{Loss Function}
The training phase requires a loss function that facilitates signatures from the same person to be close together in the embedding space, and increases the distance of signatures from different people. While contrastive loss and triplet loss work on pairs or triplets, they might not leverage information from all available negative samples effectively. To this aim, the pipeline utilizes in-batch negative loss \cite{karpukhin2020dense}, which is widely used in retrieval tasks. During training, a custom batch sampler is used to construct batches, each composed by two list of samples, a query list $B_q =\left\{X_i\right\}^{N}_{i=0}$ and a gallery list $B_g =\left\{X_j\right\}^{N}_{j=0}$, where $X_i$ are the CSI measurements and $N$ is the batch size. The i-th sample in $B_q$ and the j-th sample in $B_g$ belong to the same person if and only if $i=j$. The entire batch with both $B_q$ and $B_g$ is fed into the DNN, and consequently, the two lists of biometric signatures are computed by the model: $S_qDNN\left\{X_i\right\}^{N}_{j=0}$ $S_gDNN\left\{X_j\right\}^{N}_{j=0}$. As a result, a similarity matrix $sim(q, g)$ of size $N \times N$ is computed between the query and gallery signatures using cosine similarity. Due to the $l^2$ normalization in the Signature Module, this is simplified to the dot product:
\begin{equation}
    sim(q, g) = S_q · S^T_g.
\end{equation}

\begin{figure}[t]
    \centering
\includegraphics[width=0.75\textwidth]{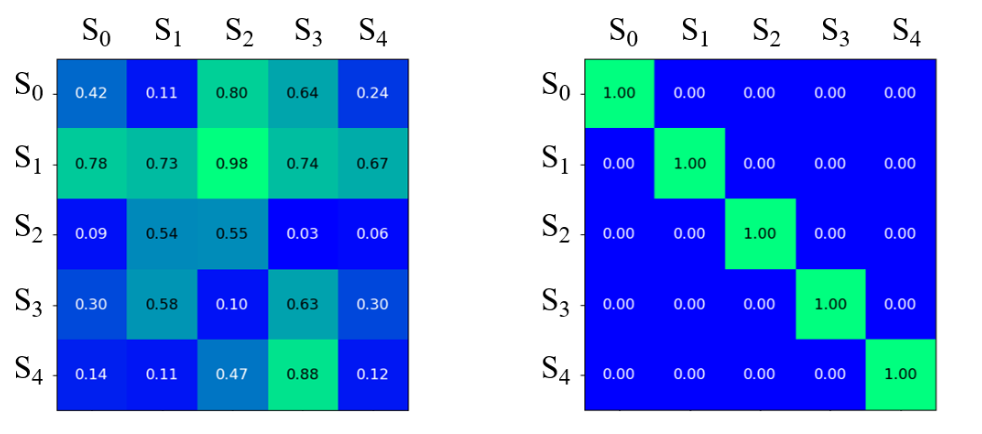}
    \caption{Similarity Matrix example used in in-batch negative loss function.}
    \label{simil}
\end{figure}

In the similarity matrix shown in Figure \ref{simil}, diagonal elements indicate similarities between each query signature and its corresponding positive gallery signature (same person), while off-diagonal elements correspond to negative pairs (different people). We apply cross-entropy loss across each row to maximize diagonal (positive) scores and minimize off-diagonal (negative) ones. For each query  $S_{q,i}$ , the softmax-normalized row is encouraged to peak at the i-th position. This leads the matrix toward an identity structure, promoting separation between individuals and clustering of same-person signatures.

\section{Experimental Results and Discussion}
\subsection{Dataset}
Experiments are conducted on the NTU-Fi dataset~\cite{dataset1,dataset2}. This dataset is created for Wi-Fi sensing applications and includes samples for both Human Activity Recognition (HAR) and Human Identification (HID). We utilize only the HID part to evaluate person Re-ID. The dataset collects the CSI measurements of 14 different subjects. For each subject, 60 samples were collected while they were performing a short walk inside the designated test area. The samples were collected in three different scenarios: subjects wearing only a T-shirt, a T-shirt and a coat, and a T-shirt, coat, and backpack, respectively. The data we recorded using two TP-Link N750 routers. The transmitter router contains a single antenna, while the receiver one contains three antennas. CSI amplitude data were collected across 114 subcarriers per antenna pair and recorded over 2000 packets per sample. As a result, each sample has a dimensionality of $3 \times 114 \times 2000$. The publicly available dataset provides only the amplitude values already extracted from the CSI, with no access to the original complex CSI matrices. The dataset is pre-divided into training and test sets, containing 546 and 294 samples respectively. To allow for evaluation during training, a 3-fold cross-validation strategy is employed, using an 80\% training and 20\% validation split within each fold.
\begin{table}[t]
\centering
\caption{Results of each model on the NTU-Fi test set.}
\label{test}
\begin{tabular}{@{}lcccc@{}}
\toprule
\textbf{Model} & \textbf{Rank-1} & \textbf{Rank-3} & \textbf{Rank-5} & \textbf{mAP} \\
\midrule
LSTM        & $0.777 \pm 0.032$ & $0.897 \pm 0.014$ & $0.933 \pm 0.005$ & $0.568 \pm 0.010$ \\
Bi-LSTM      & $0.845 \pm 0.045$ & $0.934 \pm 0.022$ & $0.958 \pm 0.013$ & $0.612 \pm 0.026$ \\
Transformer & $\mathbf{0.955 \pm 0.013}$ & $\mathbf{0.981 \pm 0.006}$ & $\mathbf{0.991 \pm 0.000}$ & $\mathbf{0.884 \pm 0.012}$ \\
\bottomrule
\end{tabular}
\end{table}
\begin{table}[t]
\centering
\caption{Performance comparison of different models with and without amplitude filtering. Metrics reported are Rank-1 accuracy and mean Average Precision (mAP). The results highlight the impact of amplitude filtering on retrieval performance across LSTM, Bi-LSTM, and Transformer-based models.}
\label{ampfilter}
\begin{tabular}{lcccc}
\toprule
& \multicolumn{2}{c}{Without filter} & \multicolumn{2}{c}{With filter} \\
\cmidrule(r){2-3} \cmidrule(r){4-5}
\textbf{Model} & \textbf{Rank-1} & \textbf{mAP} & \textbf{Rank-1} & \textbf{mAP} \\
\midrule
LSTM & 0.777 $\pm$ 0.032 & 0.568 $\pm$ 0.010 & 0.755 $\pm$ 0.038 & 0.587 $\pm$ 0.018 \\
Bi-LSTM & 0.845 $\pm$ 0.045 & 0.612 $\pm$ 0.026 & 0.786 $\pm$ 0.036 & 0.675 $\pm$ 0.018 \\
Transformers & \textbf{0.955 $\pm$ 0.013} & \textbf{0.884 $\pm$ 0.012} & 0.930 $\pm$ 0.025 & 0.851 $\pm$ 0.035 \\
\bottomrule
\end{tabular}
\end{table}
\begin{table}[t]
\centering
\caption{Effect of varying packet sizes on model performance. Results are reported for LSTM and Transformer architectures across different packet counts (100 to 2000), using Rank-1, Rank-3, Rank-5 accuracy, and mean Average Precision (mAP) as evaluation metrics. The table illustrates how performance trends shift with input granularity for both model types.}
\label{packet}
\begin{tabular}{lccccc}
\toprule
\textbf{Model} & \textbf{Packets} & \textbf{Rank-1} & \textbf{Rank-3} & \textbf{Rank-5} & \textbf{mAP} \\
\midrule
LSTM & 100  & 0.805 $\pm$ 0.050 & 0.918 $\pm$ 0.029 & 0.939 $\pm$ 0.022 & 0.597 $\pm$ 0.002 \\
LSTM & 200  & 0.777 $\pm$ 0.032 & 0.897 $\pm$ 0.014 & 0.933 $\pm$ 0.005 & 0.568 $\pm$ 0.010 \\
LSTM & 500  & 0.777 $\pm$ 0.065 & 0.906 $\pm$ 0.028 & 0.939 $\pm$ 0.017 & 0.592 $\pm$ 0.040 \\
LSTM & 1000 & 0.794 $\pm$ 0.048 & 0.991 $\pm$ 0.019 & 0.947 $\pm$ 0.011 & 0.592 $\pm$ 0.046 \\
LSTM & 2000 & 0.799 $\pm$ 0.029 & 0.915 $\pm$ 0.019 & 0.943 $\pm$ 0.013 & 0.579 $\pm$ 0.028 \\
\midrule
Transformers & 100  & 0.952 $\pm$ 0.021 & 0.983 $\pm$ 0.006 & 0.990 $\pm$ 0.005 & 0.871 $\pm$ 0.041 \\
Transformers & 200  & 0.955 $\pm$ 0.013 & 0.981 $\pm$ 0.006 & 0.991 $\pm$ 0.000 & 0.884 $\pm$ 0.012 \\
Transformers & 500  & 0.937 $\pm$ 0.020 & 0.976 $\pm$ 0.012 & 0.984 $\pm$ 0.011 & 0.840 $\pm$ 0.033 \\
Transformers & 1000 & \textbf{}0.960 $\pm$ 0.013 & 0.984 $\pm$ 0.005 & 0.988 $\pm$ 0.001 & \textbf{0.896 $\pm$ 0.020} \\
Transformers & 2000 & \textbf{0.960 $\pm$ 0.014} & 0.982 $\pm$ 0.011 & 0.990 $\pm$ 0.008 & 0.850 $\pm$ 0.054 \\
\bottomrule
\end{tabular}
\end{table}
\begin{table}[t]
\centering
\caption{Impact of data augmentation on model performance. Comparison of Rank-1 accuracy and mean Average Precision (mAP) for LSTM, BiLSTM, and Transformer models, evaluated with and without data augmentation.}
\label{aug}
\begin{tabular}{lcccc}
\toprule
& \multicolumn{2}{c}{Without augmentation} & \multicolumn{2}{c}{With augmentation} \\
\cmidrule(r){2-3} \cmidrule(r){4-5}
\textbf{Model} & \textbf{Rank-1} & \textbf{mAP} & \textbf{Rank-1} & \textbf{mAP} \\
\midrule
LSTM & 0.777 $\pm$ 0.032 & 0.568 $\pm$ 0.010 & 0.808 $\pm$ 0.038 & 0.587 $\pm$ 0.018 \\
Bi-LSTM & 0.845 $\pm$ 0.045 & 0.612 $\pm$ 0.026 & 0.889 $\pm$ 0.017 & 0.668 $\pm$ 0.016 \\
Transformers & 0.955 $\pm$ 0.013 & 0.884 $\pm$ 0.012 & 0.949 $\pm$ 0.014 & 0.860 $\pm$ 0.043 \\
\bottomrule
\end{tabular}
\end{table}
\begin{table}[t]
\centering
\caption{Evaluation of encoder type and layer depth on performance. Rank-1, Rank-3, Rank-5 accuracy, and mean Average Precision (mAP) are reported for LSTM, BiLSTM, and Transformer models with 1 and 3 encoder layers. }
\label{depth}
\begin{tabular}{lccccc}
\toprule
\textbf{Model} & \textbf{Layers} & \textbf{Rank-1} & \textbf{Rank-3} & \textbf{Rank-5} & \textbf{mAP} \\
\midrule
LSTM        & 1 & 0.777 $\pm$ 0.032 & 0.897 $\pm$ 0.014 & 0.933 $\pm$ 0.005 & 0.568 $\pm$ 0.010 \\
LSTM        & 3 & 0.822 $\pm$ 0.026 & 0.909 $\pm$ 0.004 & 0.941 $\pm$ 0.004 & 0.585 $\pm$ 0.001 \\
Bi-LSTM      & 1 & 0.845 $\pm$ 0.045 & 0.934 $\pm$ 0.022 & 0.958 $\pm$ 0.013 & 0.612 $\pm$ 0.026 \\
Bi-LSTM      & 3 & 0.825 $\pm$ 0.042 & 0.919 $\pm$ 0.012 & 0.955 $\pm$ 0.003 & 0.632 $\pm$ 0.043 \\
Transformers & 1 & \textbf{0.955 $\pm$ 0.013} & 0.981 $\pm$ 0.006 & 0.991 $\pm$ 0.000 & \textbf{0.884 $\pm$ 0.012} \\
Transformers & 3 & 0.919 $\pm$ 0.028 & 0.970 $\pm$ 0.008 & 0.984 $\pm$ 0.003 & 0.658 $\pm$ 0.026 \\
\bottomrule
\end{tabular}
\end{table}
\subsection{Implementation Details}
We train our model using an AMD Ryzen 7 processor with 8 cores (16 virtual cores), 64GB RAM and a NVIDIA GeForce RTX 3090 GPU with 24GB of RAM. For the models implementation, the Pytorch framework has been used. Regarding the training process, 300 epochs are performed for each model using a batch size of 8. Adam \cite{Adam} optimizer is used with a starting learning rate of 0.0001. A StepLR learning rate scheduler decreases the learning rate by a factor of 0.95 every 50 epochs.
\subsection{Person Re-Identification Evaluation}
To evaluate the performance our Re-ID model, mean Average Precision (mAP) has been used together with Rank-k accuracy, defined as follows:
\begin{equation}
    \text{Rank}(k) = \frac{1}{N} \sum_{i=1}^{N} \delta(r_i \le k),
\end{equation}
which provides the probability of finding the wanted subject in the top k most probable labels. The results obtained during the tests are shown in Table \ref{test}. As demonstrated, the model utilizing the Transformers encoder exceeds in performance both LSTM and Bi-LSTM ones. The Transformer-based model achieves a 95.5\% score for the Rank-1 metric and an mAP score 88.4\%. The self-attention mechanism of Transformer renders it more accurate and robust at capturing the discriminative, long-range temporal patterns within the Wi-Fi amplitude sequences relevant for Re-ID compared to the LSTM-based models. 
\subsection{Ablation Study}
Regarding amplitude filtering, Table~\ref{ampfilter} shows that models trained without the amplitude filtering pre-processing step achieved better performance. This suggests that the filtering process may have inadvertently removed useful signal variations essential for learning highly discriminative biometric signatures. As for data augmentation, Table~\ref{aug} indicates that the applied transformations improved generalization for both LSTM and Bi-LSTM architectures. In contrast, the Transformer encoder did not benefit significantly, although it consistently outperformed the other two models even without augmentation. With respect to packet size, Table~\ref{packet} reveals that LSTM performance remained mostly stable or slightly degraded with longer sequence lengths, likely due to vanishing gradient issues and limited context modeling. Conversely, the Transformer benefited from extended input sequences, thanks to its self-attention mechanism that allows efficient modeling of long-range dependencies. Only LSTM and Transformer models were evaluated in this experiment, due to the increased computational cost associated with longer inputs. Finally, we compared shallow (1-layer) and deeper (3-layer) variants of each encoder in Table~\ref{depth}. The Transformer achieved its best performance with a single layer, as deeper configurations led to overfitting and optimization instability. For LSTM and Bi-LSTM models, stacking layers resulted in marginal performance gains but introduced slower convergence and reduced training stability. These findings reinforce the overall robustness and efficiency of the Transformer encoder within the proposed framework.

\section{Conclusion}
In this paper, we presented a pipeline to address the problem of person Re-ID using Wi-Fi CSI. The proposed approach leverages a DNN that generates biometric signatures from CSI-derived features. These signatures are then compared to a gallery of known subjects to perform re-identification through similarity matching. We evaluated three encoder architectures, LSTM, Bi-LSTM, and Transformer, on the publicly available NTU-Fi dataset, with the Transformer-based model delivering the best overall performance. By applying a unified and reproducible pipeline to a public benchmark, this work establishes a valuable baseline for future research in CSI-based person re-identification. The encouraging results achieved confirm the viability of Wi-Fi signals as a robust and privacy-preserving biometric modality, and position this study as a meaningful step forward in the development of signal-based Re-ID systems.
\paragraph{Acknowledgements.}
This work was supported by the “Smart unmannEd AeRial vehiCles for Human l
\bibliographystyle{splncs04}
\bibliography{main}
\end{document}